\documentclass[10pt,twocolumn,letterpaper]{article}

\usepackage{cvpr}              
\usepackage{graphicx}
\usepackage{amsmath}
\usepackage{amssymb}
\usepackage{booktabs}
\usepackage{verbatim}
\usepackage{tablefootnote}

\usepackage[normalem]{ulem}
\usepackage{ifthen}
\usepackage{enumitem}
\usepackage{multicol}
\usepackage{multirow}

\newcommand{\cat}[0]{2016 }

\usepackage[pagebackref,breaklinks,colorlinks]{hyperref}

\usepackage[capitalize]{cleveref}
\crefname{section}{Sec.}{Secs.}
\Crefname{section}{Section}{Sections}
\Crefname{table}{Table}{Tables}
\crefname{table}{Tab.}{Tabs.}

\begin{document}

\title{CatFLW: Cat Facial Landmarks in the Wild Dataset}

\author{George Martvel \qquad Nareed Farhat \qquad Ilan Shimshoni \qquad Anna Zamansky \qquad \\
\small 
University of Haifa, Israel \qquad
}
\maketitle

\begin{abstract}

Animal affective computing is a quickly growing field of research, where only recently first efforts to go beyond animal tracking into recognizing their internal states, such as pain and emotions, have emerged. In most mammals, facial expressions are an important channel for communicating information about these states. However, 
unlike the human domain, there is an acute lack of datasets that make automation of facial analysis of animals feasible.  

This paper aims to fill this gap by presenting a dataset called Cat Facial Landmarks in the Wild (CatFLW) which contains \cat images of cat faces in different environments and conditions, annotated with 48 facial landmarks specifically chosen for their relationship with underlying musculature, and relevance to cat-specific facial Action Units (CatFACS). To the best of our knowledge, this dataset has the largest amount of cat facial landmarks available.

In addition, we describe a semi-supervised (human-in-the-loop) method of annotating images with landmarks, used for creating this dataset, which significantly reduces the annotation time and could be used for creating similar datasets for other animals.

The dataset is available on request.

\end{abstract}

\section{Introduction}
\label{sec:intro}

Emotion recognition based on facial expression and body language analysis are important and challenging topics addressed within the field of human affective computing. Comprehensive surveys cover analysis of facial expressions \cite{li2020deep}, and body behaviors \cite{noroozi2018survey}, with the recent trend being multi-modal emotion recognition approaches \cite{sharma2021survey}. Automated facial expression analysis is also studied in pain research  \cite{MAlEidan2020DeepLearningBasedMF}, and is most crucial for addressing pain in infants\cite{zamzmi2017review}.

In animals, facial expressions are known to be produced by most mammalian species\cite{mielke22facs, diogo2008fish}. Analogously to the human domain, they are assumed to convey information about emotional states, and are increasingly studied as potential indicators of subjective states in animal emotion and welfare research \cite{boneh2022explainable, merkies19eye, andresen2020towards, gleerup2015equine}. 

Although research in animal affective computing has so far lagged behind the human domain, this is rapidly changing in recent years \cite{hummel2020automatic}. 
Broome et al \cite{broome2023going} provide a comprehensive review of more than twenty studies addressing automated recognition of animals’ internal states such as emotions and pain, with the majority of these studies focusing on the analysis of animal faces in different species, such as rodents, dogs, horses and cats. 

Cats are of specific interest in the context of pain, as they are one of the most challenging species in terms of pain assessment and management due to a reduced physiological tolerance and adverse effects to common veterinary analgesics\cite{lascelles2010djd}, a lack of strong consensus over key behavioural pain indicators\cite{merola2016behavioural} and human limitations in accurately interpreting feline facial expressions\cite{dawson2019humans}. Three different manual pain assessment scales have been developed and validated in English for domestic cats: 
the UNESP-Botucatu multidimensional composite pain scale\cite{brondani2013validation}, the Glasgow composite measure pain scale (CMPS, \cite{reid2017definitive}) and the Feline Grimace Scale (FGS) \cite{evangelista2019facial}. The latter was further used for a comparative study in which human’s assignment of FGS to cats during real time observations and then subsequent FGS scoring of the same cats from still images were compared. It was shown that there was no significant difference between the scoring methods \cite{evangelista2020clinical}, indicating that facial images can be a reliable medium from which to assess pain. 

Finka and colleagues\cite{finka2019geometric} were the first to apply 48 geometric landmarks to identify and quantify facial shape change associated with pain in cats. These landmarks were based on both the anatomy of cat facial musculature, and the range of facial expressions generated as a result of facial action units (CatFACS)\cite{caeiro2017development}.  The authors manually annotated the landmarks, and used statistical methods (PCA analysis) to establish a relationship between PC scores and a well-validated measure of pain in cats. 
Feighelstein et al \cite{feighel} used the dataset and landmark annotations from \cite{finka2019geometric} to automate pain recognition in cats using machine learning models based on geometric landmarks grouped into multivectors, reaching accuracy of above 72\%, which was comparable to an alternative approach using deep learning. This indicates that the use of the 48 landmarks from \cite{finka2019geometric} can be a reliable method for pain recognition from cat facial images. The main challenge in this method remains the automation of facial landmarks detection. In general, one of the most crucial gaps in animal affective computing today, as emphasized in \cite{broome2023going}, is the lack of publically available datasets. To the best of our knowledge, all datasets containing animal faces available today have very few facial landmarks, as discussed in the next section.  

\begin{figure}[t!]
\centering
\includegraphics[width=0.4\textwidth]{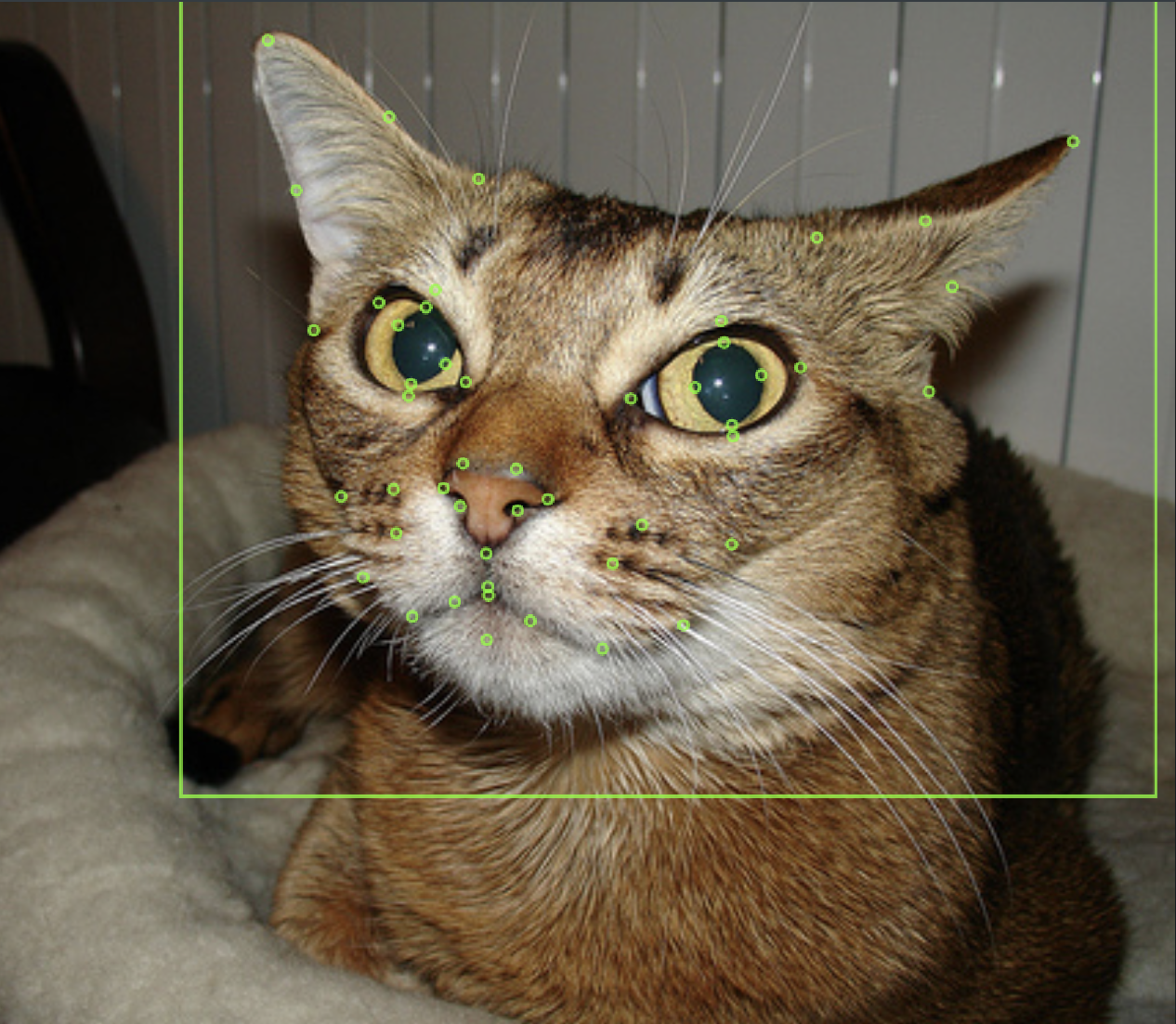}
\caption{\textbf{Annotated Cat's Face.} Image of a cat with face bounding box and 48 facial landmarks.}\label{ann_cat}\vspace{-0.25em}
\end{figure} 

To address this gap for the feline species, where facial analysis can help address acute problems in the context of pain assessment, we 
present the Cat Facial Landmarks in the Wild (CatFLW) dataset, containing images of cat faces in various conditions and environments with face bounding boxes and 48 facial landmarks of \cite{finka2019geometric,feighel} for each image (see Figure~\ref{ann_cat}). We hope that this dataset can serve as a starting point for creating similar datasets for various animals, as well as push forward automatic detection of pain and emotions in cats and other feline species. 

Our main contributions are, therefore, the following:
\begin{itemize}
    \item We present a dataset containing \cat annotated images of cats with 48 landmarks, which are based on the cat's facial anatomy and allows the study of complex morphological features of cats' faces.
    \item We describe an AI-based `human-in-the-loop' (also known as cooperative machine learning\cite{heimerl2020unraveling}) method of facial landmark annotation, and show that it significantly reduces  annotation time per image.
\end{itemize}

\section{Related Datasets}
\label{sec:datasets}

Our dataset is based on the original dataset collected by Zhang et al. \cite{original_dataset}, that contains 10,000 images of cats annotated with 9 facial landmarks. It includes a wide variety of different cat breeds in different conditions, which can provide good generalization when training computer vision models, however some images from it depict several animals, have visual interference in front of animal faces or cropped (according to our estimates 10-15\%).
In \cite{original_dataset} Zhang et al. 9 landmarks are labeled for each image, which is sufficient for detecting and analyzing general information about animal faces (for example, the tilt of the head or the direction of movement), but is not enough for analyzing complex movements of facial muscles.

\begin{table}[t!]
  \centering
  \begin{tabular}{@{}lllc@{}}
    \toprule
    Dataset & Animal & Size & Facial Landmarks \\
    \midrule
    Khan et al. \cite{AW} & Various & \textbf{21900} & 9\\
    Zhang et al. \cite{original_dataset} & Cat & 10000 & 9 \\
    Liu et al. \cite{liu2012dog} & Dog & 8351 & 8\\
    Cao et al. \cite{jinkun19cross} & Various & 5517 & 5\\
    Mougeot et al. \cite{mougeot19dogs} & Dog & 3148 & 3 \\
    Sun et al. \cite{sun20cafm} & Cat & 1706 & 15\\
    Yang et al. \cite{yang15sheep} & Sheep & 600 & 8\\
    \midrule
   CatFLW & Cat & \cat & \textbf{48}\\
    \bottomrule
  \end{tabular}
  \caption{Comparison of animal facial landmarks datasets}
  \label{tab:example}
\end{table}

Sun et al. \cite{sun20cafm} used the same dataset in their work, expanding the annotation to 15 facial landmarks.
To the best of our knowledge, only 1,706 out of the declared 10,000 images are publicly available.

Khan et al. \cite{AW} collected the AnimalWeb dataset, consisting of an impressive number of 21,900 images annotated with 9 landmarks. Despite the wide range of represented species, only approximately 450 images can be attributed to feline species.

Other relevant datasets, containing less than 9 facial landmarks, are mentioned in Table \ref{tab:example}, including those in \cite{liu2012dog, jinkun19cross, mougeot19dogs, yang15sheep}. For comparison, popular datasets for human facial landmark detection \cite{belh13lfpw, Le12helen} have several dozens of landmarks. There are also other datasets suitable for face detection and recognition of various animal species \cite{Clapham22bears, deb18monkey, guo20monkeys, korschens18elephants, chen20panda}, but they have no facial landmark annotations.

\section{The CatFLW Dataset}

\begin{figure}[t!]
\centering
\includegraphics[width=0.475\textwidth]{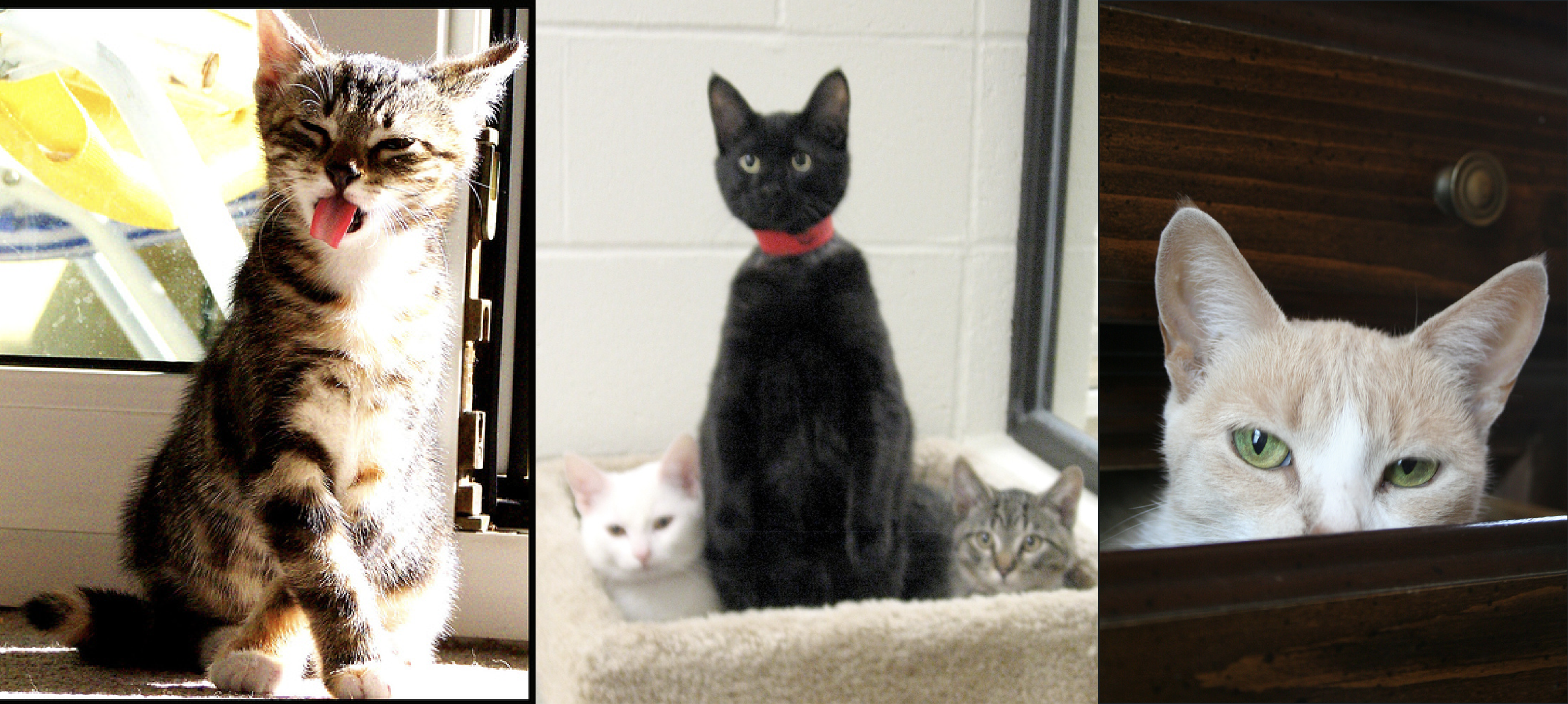}
\caption{\textbf{Examples of not suitable images from \cite{original_dataset}.} The image is considered inapplicable when it is impossible to visually determine the location of the facial landmark with full confidence or uncertainty is created due to the presence of several animals.}\label{bad_cats}\vspace{-0.75em}
\end{figure}

The CatFLW dataset consists of \cat  images selected from the dataset in \cite{original_dataset} using
the following inclusion criterion which optimize the training of landmark detection models: image should contain a single fully visible cat face, where the cat is in non-laboratory conditions (`in the wild'). Thus many images from the original dataset were discarded (Figure~\ref{bad_cats} shows the examples).

\begin{figure}[b!]
\centering
\hspace*{-0.3cm} 
\includegraphics[width=0.5\textwidth]{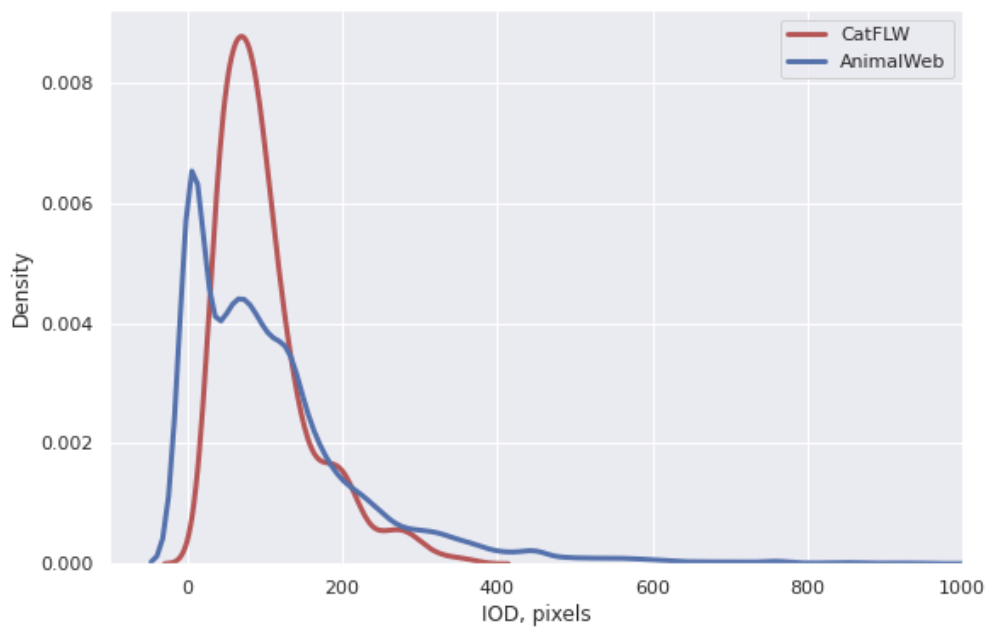}
\caption{\textbf{IOD distributions for CatFLW and AnimalWeb.} Both datasets have approximately the same mean inter-ocular distance, however, the latter has both a high number of small-sized faces and a wide variation in large faces.}\label{comp}
\end{figure}

The images in this dataset have a large variation of colors, breeds and environments, as well as of scale, position and angle of head tilts. Figure \ref{comp} shows the distribution of inter-ocular distances (IOD --- distance between the two medial canthi of each eye) on CatFLW and  AnimalWeb \cite{AW}. CatFLW dataset has similar characteristics to AnimalWeb, except that the latter has about twice the standard deviation. This can be explained by the large species diversity, as well as a greater variation in the size of the images. 

\subsection{The Landmark Annotation Process}

After selection and filtering of images, a bounding box and 48 facial landmarks were placed on each image using the Labelbox platform\cite{labelbox}.

{\bf Bounding box.} In the current dataset, the bounding box is not tied to facial landmarks (this is usually done by selecting the outermost landmarks and calculating the difference in their coordinates), but is visually placed so that the entire face of the animal fits into the bounding box as well as about 10\% of the space around the face. This margin is made because when studying detection models on the initial version of the dataset, it was noticed that some of them tend to crop the faces and choose smaller bounding boxes. In the case when face detection is performed for the further localization of facial landmarks, the clipping of faces can lead to the disappearance from the image of important parts of the cat's face, such as the tips of the ears or the mouth. Such an annotation still makes it possible to construct bounding boxes by outermost landmarks if needed.

{\bf Landmarks.} 48 facial landmarks introduced in Finka et al. \cite{finka2019geometric} and used in Feighelstein et al. \cite{feighel} were manually placed on each image. These landmarks were specifically chosen for their relationship with underlying musculature and anatomical features, and relevance to cat-specific Facial Action Units (CatFACS) \cite{caeiro2017development}.

{\bf Annotation Process.} The process of landmark annotation had several stages: at first, 10\% images were annotated by an annotator who has an extensive experience labeling facial landmarks for various animals, then they were annotated by second expert with the same annotation instructions. The landmarks were then compared to verify the internal validity and reliability via the Inter Class Correlation Coefficient ICC2 \cite{shrout1979intraclass}, and reached a strong agreement between the annotators with a score of 0.998.

Finally, the remaining part of images were annotated by the first annotator using the ‘human-in-the-loop' method described below.

After the annotation, a review and correction of landmarks and bounding boxes were performed, all images were re-filtered and verified.

\begin{figure*}[t!]
\centering
\includegraphics[width=1\textwidth]{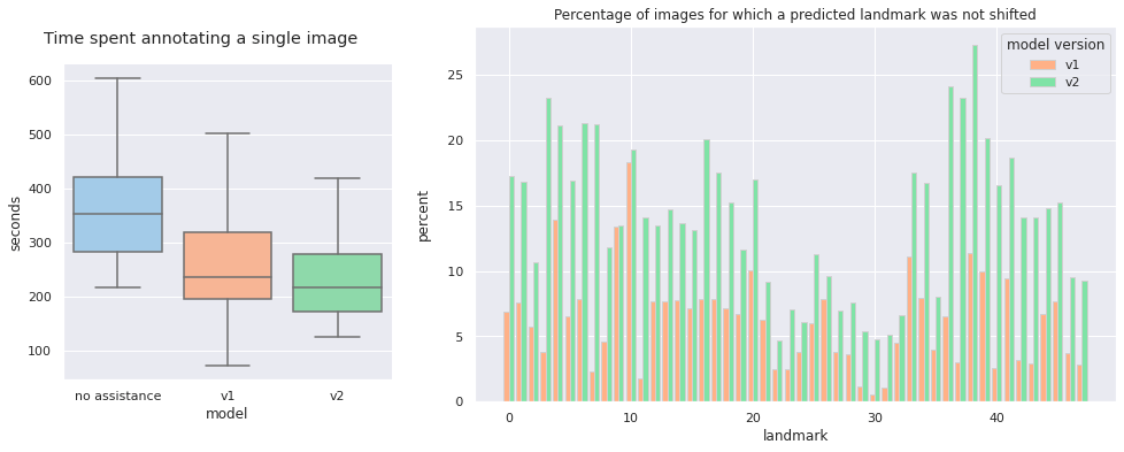}
\caption{\textbf{Left: The distribution of annotation time per image for different batches.} The predictions of the first model significantly reduced the annotation time, while the subsequent refinement by the second model reduced the variance, without changing the median value too much. \textbf{Right: The percentage of images from the batch for which a predicted landmark was not shifted.} The more accurate the model's predictions are, the less time is needed for annotation: correctly predicted landmarks do not need to be shifted.}\label{bars}\vspace{-1em}
\end{figure*}

\subsection{AI-assisted Annotation}
 
Khan et al.\cite{AW} spent approximately 5,408 man-hours annotating all the images in the AnimalWeb dataset (each annotation was obtained by taking the median value of the annotations of five or more error-prone volunteers). Roughly, it is 3 minutes for the annotation of one image (9 landmarks) or 20 seconds for one landmark (taken per person). Our annotation process took $\sim$140 hours, which is about 4.16 minutes per image (48 landmarks) and only 5.2 seconds per landmark. Such performance is achieved due to a semi-supervised 'human-in-the-loop’  method that uses the predictions of a gradually-trained model as a basis for annotation.

To assess the impact of AI on the annotation process, 3 batches were created from the data. The first one consisted of two hundred images and was annotated without AI assistance. Next, we trained the EfficientNet-based \cite{efnet} model on the dataset from \cite{finka2019geometric} (about 600 images) combined with the first annotated batch ($\sim$800 images total). After that, we used this model (v1) to predict the landmarks on the second batch ($\sim$900 images). Based on the predictions, landmarks were annotated for the second batch. This approach made it possible to reduce the average time for annotation of one image between the first and second batches by 35\%, since for annotation it became necessary only to adjust the position of the landmark, and not to place it from scratch. However, the predictions obtained turned out to be inaccurate, so the first and second batches were combined to train the next version of the model (v2) with the same architecture ($\sim$1700 images total). After training, the v2 model was used to predict the landmarks on the third batch, and then an annotation was performed based on these predictions. The results obtained for the time spent on annotation of one image are shown in Figure \ref{bars}(left). Increasing the accuracy of predicting the position of a landmark reduces the time required to adjust it, sometimes to zero if the annotator considers the position to be completely correct. Figure \ref{bars}(right) shows the distribution of the percentage of images in which a specific facial landmark was not shifted by the annotator: between the two versions of the model, the increase is approximately doubled.

\section{Conclusion}

In this paper we present a CatFLW dataset of annotated cat faces. It includes \cat images of cats in various conditions with face bounding boxes and 48 facial landmarks for each cat. Each of the landmarks corresponds to a certain anatomical feature, which makes it possible to use these landmarks for various kinds of tasks on evaluating the internal state of cats. The proposed AI-assisted method of image annotation has significantly reduced the time for annotation of landmarks and reduced the amount of manual work during the preparation of the dataset. The created dataset opens up opportunities for the development of various computer vision models for detecting facial landmarks of cats and can be the starting point for research in the field of behaviour observation and well-being of cats and animals in general.

\section*{Acknowledgements}

The research was supported by the Data Science Research
Center at the University of Haifa. We thank Ephantus Kanyugi for his contribution with data annotation and management. We thank Yaron Yossef for his technical support. 

{\small
\bibliographystyle{ieee_fullname}
\bibliography{egbib}

\begin{thebibliography}{10}\itemsep=-1pt

\bibitem{labelbox}
Labelbox, "labelbox," online, 2023. [online]. available:.
\newblock \url{https://labelbox.com}.

\bibitem{MAlEidan2020DeepLearningBasedMF}
Rasha~M Al-Eidan, Hend~Suliman Al-Khalifa, and AbdulMalik~S Al-Salman.
\newblock Deep-learning-based models for pain recognition: A systematic review.
\newblock {\em Applied Sciences}, 10:5984, 2020.

\bibitem{andresen2020towards}
Niek Andresen, Manuel W{\"o}llhaf, Katharina Hohlbaum, Lars Lewejohann, Olaf
  Hellwich, Christa Th{\"o}ne-Reineke, and Vitaly Belik.
\newblock Towards a fully automated surveillance of well-being status in
  laboratory mice using deep learning: Starting with facial expression
  analysis.
\newblock {\em PLoS One}, 15(4):e0228059, 2020.

\bibitem{belh13lfpw}
Peter~N Belhumeur, David~W Jacobs, David~J Kriegman, and Neeraj Kumar.
\newblock Localizing parts of faces using a consensus of exemplars.
\newblock {\em IEEE transactions on pattern analysis and machine intelligence},
  2013.

\bibitem{boneh2022explainable}
Tali Boneh-Shitrit, Marcelo Feighelstein, Annika Bremhorst, Shir Amir, Tomer
  Distelfeld, Yaniv Dassa, Sharon Yaroshetsky, Stefanie Riemer, Ilan Shimshoni,
  Daniel~S Mills, et~al.
\newblock Explainable automated recognition of emotional states from canine
  facial expressions: the case of positive anticipation and frustration.
\newblock {\em Scientific reports}, 12(1):22611, 2022.

\bibitem{brondani2013validation}
Juliana~T Brondani, Khursheed~R Mama, Stelio~PL Luna, Bonnie~D Wright, Sirirat
  Niyom, Jennifer Ambrosio, Pamela~R Vogel, and Carlos~R Padovani.
\newblock Validation of the english version of the unesp-botucatu
  multidimensional composite pain scale for assessing postoperative pain in
  cats.
\newblock {\em BMC Veterinary Research}, 9(1):1--15, 2013.

\bibitem{broome2023going}
Sofia Broome, Marcelo Feighelstein, Anna Zamansky, Carreira~G Lencioni,
  Haubro~P Andersen, Francisca Pessanha, Marwa Mahmoud, Hedvig Kjellstr{\"o}m,
  and Albert~Ali Salah.
\newblock Going deeper than tracking: A survey of computer-vision based
  recognition of animal pain and emotions.
\newblock {\em International Journal of Computer Vision}, 131(2):572--590,
  2023.

\bibitem{caeiro2017development}
C{\'a}tia~C Caeiro, Anne~M Burrows, and Bridget~M Waller.
\newblock Development and application of catfacs: Are human cat adopters
  influenced by cat facial expressions?
\newblock {\em Applied Animal Behaviour Science}, 2017.

\bibitem{jinkun19cross}
Jinkun Cao, Hongyang Tang, Haoshu Fang, Xiaoyong Shen, Cewu Lu, and Yu{-}Wing
  Tai.
\newblock Cross-domain adaptation for animal pose estimation.
\newblock {\em CoRR}, abs/1908.05806, 2019.

\bibitem{chen20panda}
Peng Chen, Pranjal Swarup, Wojciech~M Matkowski, Kong~Adams WK, Su Han, Zhihe
  Zhang, and Hou Rong.
\newblock A study on giant panda recognition based on images of a large
  proportion of captive pandas.
\newblock {\em Ecol Evol.}, 2020.

\bibitem{Clapham22bears}
Melanie Clapham, Ed Miller, Mary Nguyen, and Russel~C Van~Horn.
\newblock Multispecies facial detection for individual identification of
  wildlife: a case study across ursids.
\newblock {\em Mamm Biol 102, 943–955}, 2022.

\bibitem{dawson2019humans}
Lauren Dawson, Joanna Cheal, Lauren Niel, and Georgia Mason.
\newblock Humans can identify cats’ affective states from subtle facial
  expressions.
\newblock {\em Animal Welfare}, 28(4):519--531, 2019.

\bibitem{deb18monkey}
Debayan Deb, Susan Wiper, Sixue Gong, Yichun Shi, Cori Tymoszek, Alison
  Fletcher, and Anil~K. Jain.
\newblock Face recognition: Primates in the wild.
\newblock In {\em 2018 IEEE 9th International Conference on Biometrics Theory,
  Applications and Systems (BTAS)}, pages 1--10, 2018.

\bibitem{diogo2008fish}
Rui Diogo, Virginia Abdala, N Lonergan, and BA Wood.
\newblock From fish to modern humans--comparative anatomy, homologies and
  evolution of the head and neck musculature.
\newblock {\em Journal of Anatomy}, 213(4):391--424, 2008.

\bibitem{evangelista2020clinical}
Marina~C Evangelista, Javier Benito, Beatriz~P Monteiro, Ryota Watanabe,
  Graeme~M Doodnaught, Daniel~SJ Pang, and Paulo~V Steagall.
\newblock Clinical applicability of the feline grimace scale: real-time versus
  image scoring and the influence of sedation and surgery.
\newblock {\em PeerJ}, 8:e8967, 2020.

\bibitem{evangelista2019facial}
Marina~C Evangelista, Ryota Watanabe, Vivian~SY Leung, Beatriz~P Monteiro,
  Elizabeth O’Toole, Daniel~SJ Pang, and Paulo~V Steagall.
\newblock Facial expressions of pain in cats: the development and validation of
  a feline grimace scale.
\newblock {\em Scientific reports}, 9(1):1--11, 2019.

\bibitem{feighel}
Marcelo Feighelstein, Ilan Shimshoni, Lauren Finka, Stelio~P. Luna, Daniel
  Mills, and Anna Zamansky.
\newblock Automated recognition of pain in cats.
\newblock {\em Scientific Reports}, 12, 2022.

\bibitem{finka2019geometric}
Lauren~R Finka, Stelio~P Luna, Juliana~T Brondani, Yorgos Tzimiropoulos, John
  McDonagh, Mark~J Farnworth, Marcello Ruta, and Daniel~S Mills.
\newblock Geometric morphometrics for the study of facial expressions in
  non-human animals, using the domestic cat as an exemplar.
\newblock {\em Scientific reports}, 9(1):1--12, 2019.

\bibitem{gleerup2015equine}
Karina~B Gleerup, Bj{\"o}rn Forkman, Casper Lindegaard, and Pia~H Andersen.
\newblock An equine pain face.
\newblock {\em Veterinary anaesthesia and analgesia}, 42(1):103--114, 2015.

\bibitem{guo20monkeys}
Songtao Guo, Pengfei Xu, Qiguang Miao, Guofan Shao, Colin~A Chapman, Xiaojiang
  Chen, Gang He, Dingyi Fang, He Zhang, Yewen Sun, Zhihui Shi, and Baoguo Li.
\newblock Automatic identification of individual primates with deep learning
  techniques.
\newblock {\em iScience}, 2020.

\bibitem{heimerl2020unraveling}
Alexander Heimerl, Katharina Weitz, Tobias Baur, and Elisabeth Andr{\'e}.
\newblock Unraveling ml models of emotion with nova: Multi-level explainable ai
  for non-experts.
\newblock {\em IEEE Transactions on Affective Computing}, 13(3):1155--1167,
  2020.

\bibitem{hummel2020automatic}
Hilde~I Hummel, Francisca Pessanha, Albert~Ali Salah, Thijs~JPAM van Loon, and
  Remco~C Veltkamp.
\newblock Automatic pain detection on horse and donkey faces.
\newblock In {\em FG}, 2020.

\bibitem{AW}
Muhammad~Haris Khan, John McDonagh, Salman~H Khan, Muhammad Shahabuddin, Aditya
  Arora, Fahad~Shahbaz Khan, Ling Shao, and Georgios Tzimiropoulos.
\newblock Animalweb: {A} large-scale hierarchical dataset of annotated animal
  faces.
\newblock {\em CVRR}, abs/1909.04951, 2019.

\bibitem{korschens18elephants}
Matthias K{\"{o}}rschens, Bj{\"{o}}rn Barz, and Joachim Denzler.
\newblock Towards automatic identification of elephants in the wild.
\newblock {\em CoRR}, abs/1812.04418, 2018.

\bibitem{lascelles2010djd}
B~Duncan~X Lascelles and Sheilah~A Robertson.
\newblock Djd-associated pain in cats: what can we do to promote patient
  comfort?
\newblock {\em Journal of Feline Medicine \& Surgery}, 12(3):200--212, 2010.

\bibitem{Le12helen}
Vuong Le, Jonathan Brandt, Zhe Lin, Lubomir Bourdev, and Thomas~S Huang.
\newblock Interactive facial feature localization.
\newblock {\em Proceedings of European Conference on Computer Vision (ECCV),
  Springer}, 2012.

\bibitem{li2020deep}
Shan Li and Weihong Deng.
\newblock Deep facial expression recognition: A survey.
\newblock {\em IEEE transactions on affective computing}, 13(3):1195--1215,
  2022.

\bibitem{liu2012dog}
Jiongxin Liu, Angjoo Kanazawa, David Jacobs, and Peter Belhumeur.
\newblock Dog breed classification using part localization.
\newblock In {\em European conference on computer vision}, pages 172--185.
  Springer, 2012.

\bibitem{merkies19eye}
Katrina Merkies, Chloe Ready, Leanne Farkas, and Abigail Hodder.
\newblock Eye blink rates and eyelid twitches as a non-invasive measure of
  stress in the domestic horse.
\newblock {\em Animals (Basel)}, 2019.

\bibitem{merola2016behavioural}
Isabella Merola and Daniel~S Mills.
\newblock Behavioural signs of pain in cats: an expert consensus.
\newblock {\em PloS one}, 11(2):e0150040, 2016.

\bibitem{mielke22facs}
Alexander Mielke, Bridget~M Waller, Claire Pérez, Alan~V Rincon, Julie
  Duboscq, and Jérôme Micheletta.
\newblock Netfacs: Using network science to understand facial communication
  systems.
\newblock {\em Behavior Research Methods 54, 1912–1927}, 2022.

\bibitem{mougeot19dogs}
Guillaume Mougeot, Dewei Li, and Shuai Jia.
\newblock A deep learning approach for dog face verification and recognition.
\newblock {\em Lecture Notes in Computer Science}, 2019.

\bibitem{noroozi2018survey}
Fatemeh Noroozi, Ciprian~Adrian Corneanu, Dorota Kami{\'n}ska, Tomasz
  Sapi{\'n}ski, Sergio Escalera, and Gholamreza Anbarjafari.
\newblock Survey on emotional body gesture recognition.
\newblock {\em IEEE transactions on affective computing}, 12(2):505--523, 2018.

\bibitem{reid2017definitive}
Jacky Reid, Mariann~E Scott, Gillian Calvo, and Andrea~M Nolan.
\newblock Definitive glasgow acute pain scale for cats: validation and
  intervention level.
\newblock {\em Veterinary Record}, 180(18):449--449, 2017.

\bibitem{sharma2021survey}
Garima Sharma and Abhinav Dhall.
\newblock A survey on automatic multimodal emotion recognition in the wild.
\newblock In Gloria Phillips-Wren, Anna Esposito, and Lakhmi~C. Jain, editors,
  {\em Advances in Data Science: Methodologies and Applications}, pages 35--64.
  Springer International Publishing, Cham, 2021.

\bibitem{shrout1979intraclass}
Patrick~E Shrout and Joseph~L Fleiss.
\newblock Intraclass correlations: Uses in assessing rater reliability.
\newblock {\em Psychological Bulletin}, 86(2):420--428, 1979.

\bibitem{sun20cafm}
Yifan Sun and Noboru Murata.
\newblock Cafm: A 3d morphable model for animals.
\newblock {\em IEEE Winter Applications of Computer Vision Workshops (WACVW)},
  2020.

\bibitem{efnet}
Mingxing Tan and Quoc~V Le.
\newblock Efficientnet: Rethinking model scaling for convolutional neural
  networks.
\newblock {\em CoRR}, abs/1905.11946, 2019.

\bibitem{yang15sheep}
Heng Yang, Renqiao Zhang, and Peter Robinson.
\newblock Human and sheep facial landmarks localisation by triplet interpolated
  features.
\newblock {\em CVRR}, abs/1509.04954, 2015.

\bibitem{zamzmi2017review}
Ghada Zamzmi, Rangachar Kasturi, Dmitry Goldgof, Ruicong Zhi, Terri Ashmeade,
  and Yu Sun.
\newblock A review of automated pain assessment in infants: features,
  classification tasks, and databases.
\newblock {\em IEEE reviews in biomedical engineering}, 11:77--96, 2017.

\bibitem{original_dataset}
Weiwei Zhang, Jian Sun, and Xiaoou Tang.
\newblock Cat head detection - how to effectively exploit shape and texture
  features.
\newblock {\em ECCV}, 2008.

\end{thebibliography}
}

\end{document}